\documentclass{article}

\usepackage{arxiv}

\usepackage[utf8]{inputenc} 
\usepackage[T1]{fontenc}    
\usepackage{hyperref}       
\usepackage{url}            
\usepackage{booktabs}       
\usepackage{amsfonts}       
\usepackage{nicefrac}       
\usepackage{microtype}      
\usepackage{lipsum}		
\usepackage{graphicx}
\usepackage{natbib}
\usepackage{doi}
\usepackage{amsmath}
\usepackage{cleveref}

\usepackage{subcaption}
\usepackage{xcolor}
\usepackage{hyperref}
\usepackage{booktabs, subcaption, colortbl,xcolor}

\title{Retrieval-Augmented Personalization with Foundation Models for Wearable Stress Detection}

\author{ {Louis Simon}\\
	Institut des Systèmes Intelligents et de Robotique\\
	  Sorbonne University\\
	Paris, France \\
	\texttt{louis.simon@isir.upmc.fr} \\
	\And
	{Mohamed Chetouani}\\
	Institut des Systèmes Intelligents et de Robotique\\
	  Sorbonne University\\
	Paris, France \\
	\texttt{mohamed.chetouani@isir.upmc.fr} \\
}

\renewcommand{\shorttitle}{Retrieval-Augmented Personalization with Foundation Models for Wearable Stress Detection}

\date{}
\hypersetup{
pdftitle={A template for the arxiv style},
pdfsubject={q-bio.NC, q-bio.QM},
pdfauthor={David S.~Hippocampus, Elias D.~Striatum},
pdfkeywords={First keyword, Second keyword, More},
}

\begin{document}
\maketitle
\vspace{.5cm}

\begin{abstract}
Personalization in wearable-based stress detection remains challenging due to substantial inter-individual variability in physiological and behavioral responses. While traditional approaches rely on user-specific fine-tuning or costly self-supervised pre-training on large datasets, we propose a lightweight alternative based on retrieval-augmented personalization. Our method leverages frozen, out-of-domain foundation models to retrieve similar patterns from a target user's history and encode them into a compact personalized embedding that modulates representations extracted by a lightweight transformer network. We evaluate our approach on the WESAD stress detection dataset with N=15 users, comprising wrist-worn physiological (EDA, BVP, temperature) and activity (accelerometer) signals, and report gains of +3.92\% in accuracy and +4.76\% in macro F1-score over a non-personalized transformer baseline, approaching supervised fine-tuning performance without requiring any labeled user data. We further show that temporal retrieval, where only prior user samples are available, achieves performance close to full intra-user retrieval, demonstrating robustness to limited user history. Finally, we explore personalization in a cross-dataset retrieval setting, leveraging embeddings from the K-Emocon dataset to personalize representations for stress detection on the WESAD dataset.
\end{abstract}


\section{Introduction}
Personalization refers to the process of tailoring a model's output to an individual's profile, preferences, or behaviors. In the context of affective computing, personalization pertains to accounting for individual behavioral differences that exist when expressing emotions or experiencing stress \cite{li_survey_2023-1, han_systematic_2024}. Such individual differences can stem from various factors, e.g., age, gender, culture, personality, but also from behavioral or physiological differences, that are much more complex to unveil. Personalization techniques in Affective Computing often leverage small subsets of labeled data from target users to train, fine-tune  or select models \cite{behinaein_transformer_2021, taylor_personalized_2020, yu_personalized_2021}. While efficient, such approaches require labeled user data, which may be unavailable in some scenarios. With the advent of transformers and foundation models, more researchers are shifting towards generalized models to tackle datasets with high inter-individual variability \cite{erturk_beyond_2025, pillai_papagei_2024, yuan_self-supervised_2024}. Such approach builds on largely pre-trained models' ability to represent patterns and capture invariance to reduce the variability induced by individual differences. Several studies have demonstrated the efficacy of self-supervised pre-training on physiological data on various affective computing downstream tasks \cite{wu_transformer-based_2024, dissanayake_sigrep_2022}. Pre-training such model can, however, be costly, requiring large datasets, which are still small compared to modern image or text pretraining dataset. Finally, studies often fail to report meaningful metrics of personalization. In particular, when it comes to evaluating individual users, most of the metrics report average performance across all users.
In this work, we propose methods that leverage learned representations from foundation models. Numerous studies have used conditioning on user demographics, which can raise ethical questions, or score, either from clinical or personality tests, to personalize model \cite{rudovic_personalized_2018, yang_attribute-invariant_2019}. Acquiring such information could be tedious and the knowledge of the relation between high level individual traits and physiological responses is still unknown. Therefore, we focus on exploiting abstract representations from a generalist foundation model to condition our model for personalization. Such techniques have been extensively studied in domains such as recommendation systems \cite{dhelim_survey_2022, he_survey_2023}, personalized chatbots \cite{zhang_personalization_2025}, or healthcare \cite{liang_survey_2019}. We draw inspiration from recent work on personalized LLM \cite{doddapaneni_user_2024, ning_user-llm_2025}, and adapt it to wearable-based stress detection. Our contributions are as follows: (1) a personalized model trained end-to-end that conditions predictions on a user representation, (2) a personalization module that builds this representation by retrieving similar patterns from the user's own history using a frozen pre-trained model, requiring no labeled user data, (3) a study of cold-start personalization, evaluating how our method behaves as the user's history grows over time, and (4) insights on the role of pre-trained models as retrieval backbones.

\section{RELATED WORKS}
\subsection{Personalization in Affective Computing}

Personalization is a key challenge when working with human behavioral data, as subjects usually exhibit high inter-individual variability in both physiological responses and emotional expression. Following Li et al. survey \cite{li_survey_2023-1}, personalization techniques in affective computing can be broadly split into two approaches: data-based, which involve selecting, grouping, or modifying data prior to training, and model-based, which consist of specific training procedures designed to account for individual differences.

The most straightforward approach consists in training one model per individual \cite{han_systematic_2024, cao_speaker-sensitive_2015} or groups \cite{salam_learning_2022}, as opposed to a single population-level model. This is however rarely feasible for deep learning models, which require substantial amounts of data per user. A natural relaxation is to share a common feature encoder while training separate classification heads per user, i.e., multitask learning \cite{taylor_personalized_2020, saeed_model_2018}. When a small amount of labeled data is available for new users, fine-tuning offers an efficient way to adapt a generalized model, although it is prone to overfitting with very few samples \cite{han_systematic_2024, behinaein_transformer_2021}. A principled solution to this low-data regime is meta-learning, where users are treated as tasks and the objective is to learn an initialization that quickly adapts to new ones \cite{yu_personalized_2021, kambale_analyzing_2024}. However, all of the above approaches require at least some labeled data per user, which can be difficult and costly to acquire in practice.

An alternative line of work avoids per-sample annotations entirely by conditioning models on user attributes or personal information, allowing the model to learn the underlying correlation between signals and user characteristics. For instance, Yang and Lee \cite{yang_attribute-invariant_2019} condition a variational autoencoder on Big Five personality scores to learn attribute-invariant representations of physiological signals for emotion recognition. In a more advanced approach, Gerczuk et al. \cite{gerczuk_zero-shot_2023} proposed hypernetworks that generate model parameters conditioned on user metadata, enabling zero-shot adaptation to new users without requiring any of their signal data. In a study on personalized robot interaction with children with ASD, Rudovic et al. \cite{rudovic_culturenet_2018} combine conditioning on a clinical severity score (CARS) with a multi-branch architecture that captures group-level variation across culture, gender, and individual, demonstrating the benefit of integrating both user attributes and structured population priors. However, such attributes are often difficult to acquire, requiring either self-report questionnaires or clinical diagnosis, and may introduce bias when demographic information is used as a proxy for individual physiological differences. More broadly, these methods depend on the availability of user-specific side information that may be impractical to collect in passive sensing scenarios.

One way to avoid any dependence on personal information and annotation builds explicit user representations from the data only. For instance, Barros et al. \cite{barros_adapting_2022} proposed an affective memory system that encodes user-specific responses to stimuli as slots in a dynamic memory structure, updated in a continual fashion as the user interacts with the system. Triantafyllopoulos et al. \cite{triantafyllopoulos_enrolment-based_2024} proposed to encode a small set of annotated speech samples into a user embedding that conditions a speech emotion recognition model. While their approach leverage annotated samples, they are connected at enrollment, i.e., only once and by user themselves, therefore enabling adaptation with minimal user-specific information and without full model retraining. Most closely related to our work, Tran et al. \cite{tran_personalized_2023} learn a user embedding by jointly optimizing a supervised emotion loss regularized by a HuBERT self-supervised objective, and perform label distribution calibration by retrieving users with similar embeddings in the HuBERT latent space. While sharing our intuition of leveraging pretrained representations for user-aware inference, their approach retrieves similar \textit{users} rather than similar \textit{samples}, requires labeled emotion annotations to train the user embedding, and operates on speech with a domain-specific pretrained model. 

In contrast, our method constructs a personalization embedding from unlabeled physiological recordings via frozen, out-of-domain foundation models, requiring no affect annotations and no physiological pretraining, making it well-suited to passive wearable sensing scenarios where rich unlabeled user history is available but annotation is impractical.

\subsection{Foundation models for wearable sensing}
The rise of self-supervised transformers has enabled the emergence of generalizable representations that transfer across tasks and domains. In the context of affective computing and personal sensing, several studies have demonstrated that large pretrained models can serve as effective feature extractors for wearable data without task-specific pretraining, either by using it as input features or by direct linear probing of the foundation models \cite{schuller_affective_2026, lian_merbench_2026, zhao_multi-level_2022}. A parallel line of work has explored training domain-specific SSL models directly on physiological signals. For instance, Wu et al. \cite{wu_transformer-based_2024} pretrained a multimodal transformer on a closed-source dataset of 4M samples using a signal transformation pretext task, achieving strong performance on WESAD and K-EmoCon. Similarly in \cite{dissanayake_sigrep_2022}, authors pretrained a transformer model on a collection of wearable physiology datasets. While effective, such approaches require substantial computational resources and large-scale physiological datasets that remain scarce compared to vision or text domains.

More recently, foundation models specifically designed for wearable and physiological data have begun to emerge. PaPaGei\cite{pillai_papagei_2024} offers pretrained representations for optical physiological signals, while Erturk et al. \cite{erturk_beyond_2025} demonstrated the value of large-scale behavioral wearable data for health prediction. However, these models typically assume fixed input modalities or formats, PPG-only inputs, fixed channel configurations, or patch-based tokenization, limiting their applicability across heterogeneous wearable setups. Furthermore, efficient fine-tuning strategies such as LoRA, while promising for adapting large language models, remain difficult to apply in wearable affective computing due to the absence of sufficiently large physiological foundation models and the limited dataset sizes typical of this domain.

An alternative is to leverage out-of-domain foundation models whose representations transfer to physiological signals without any physiological pretraining. Two recent studies have shown that speech foundation models can extract meaningful representations from physiological time series \cite{phukan_beyond_2024, ahmed_al_dossary_speech_2025}, achieving strong performance despite the apparent domain mismatch, though the reasons for this transferability remain an open question. Building on this line of work, we extend the use of out-of-domain foundation models to time series and speech foundation models, and evaluate both families as retrieval backbones for personalized physiological affect recognition. Crucially, we use these models in a fully frozen setting, requiring no physiological data for pretraining and supporting flexible multimodal input by processing each modality independently.

\begin{figure*}[]
    \centering
    \includegraphics[width=\linewidth]{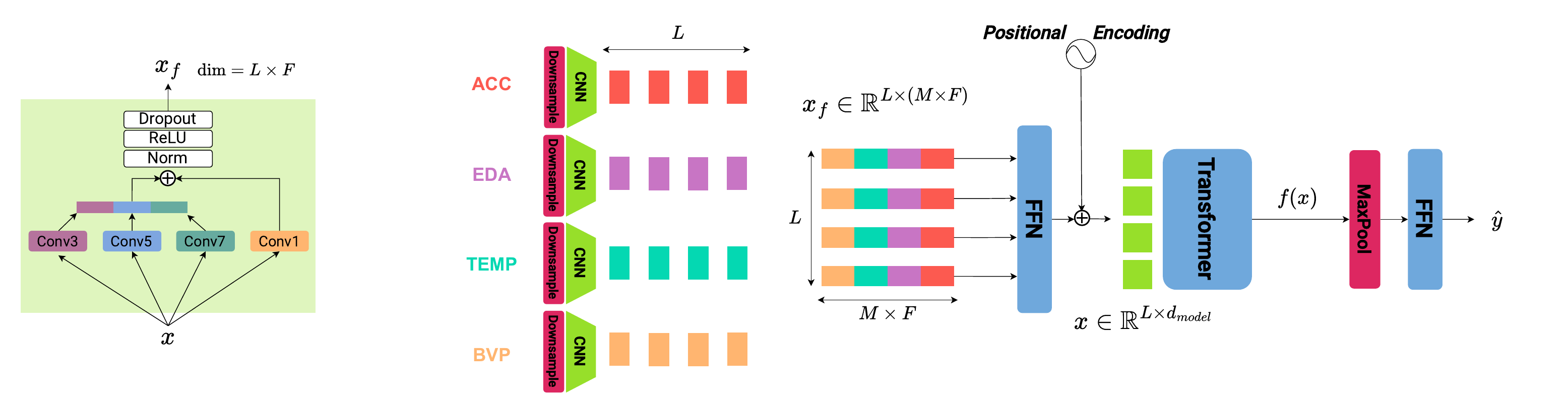}
    \caption{\textbf{CNN + Transformer Backbone}: Wearable input signals are processed individually with Inception-like CNNs (left), fused and jointly process by a Transformer. The full architecture with max pooling and a prediction (right) serves as a baseline.}
    \label{fig:transformer_archi}
\end{figure*} 


\section{Methods}
\begin{figure}[h!]
    \centering
    \includegraphics[width=0.7\linewidth]{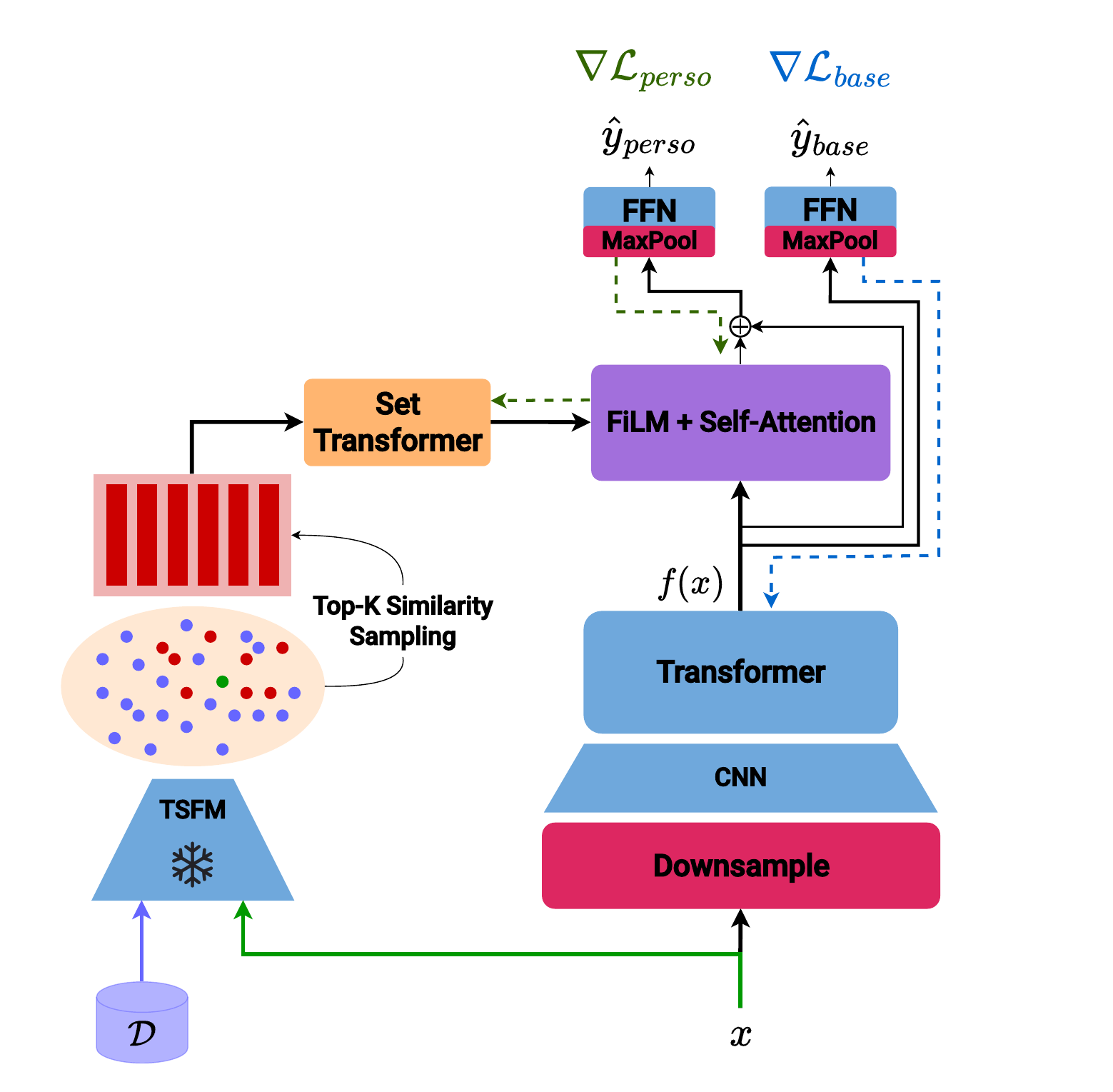}
    \caption{\textbf{Our proposed personalized model} conditions non-personalized representation $f(x)$ with intra-user contextual representations built via foundation model embedding retrieval.}
    \label{fig:full_archi}
\end{figure}

In this work, we draw inspiration from UserLLM \cite{ning_user-llm_2025}, which uses a user embedding module and interleaved cross-attention to condition a frozen LLM on learned user embeddings. Adapting this to wearable-based stress detection faces three challenges: (1) the absence of large-pretrained foundation models for wearable with flexible input modalities, (2) limited data availability and high inter-individual heterogeneity that prevent training complex deep architectures, and (3) the smaller number of users in the WESAD dataset ($N = 15$) compared to MovieLens ($N = 138$ k in UserLLM), making user embedding pre-training impractical. Instead, we propose to train the predictive branch, analogous to the LLM in UserLLM, from scratch as a lightweight transformer, while leveraging a general-purpose out-of-domain Foundation Model (FM) to build contextualized user representations through similarity-based pattern retrieval, as depicted in \Cref{fig:full_archi}. Given an input multimodal signal $x$, we extract non-personalized representations $f(x)$ from the transformer encoder, retrieve similar user patterns in the FM embedding space, and pool them into a compact embedding that modulates $f(x)$ via FiLM layers. In the following sections, we describe in detail each design steps of our approach, emphasizing on its relevance for small datasets with high inter-individual differences.

\subsection{Encoder}
Transformers have recently shown great potential for time series and physiological signals. However, it can be difficult to efficiently train them with a limited amount of data, especially without pre-training. Therefore, we adopt a rather simple and lightweight architecture inspired by prior work \cite{dissanayake_sigrep_2022}. Each of the $M$ modalities are first down-sampled and processed by a CNN encoder built with inception-like blocks. After modality-wise encoding, convolutional features are then concatenated feature-wise, projected with a single layer feedforward neural network and passed to a shallow Transformer. This CNN-Transformer backbone produces a non-personalized representation $f(x)$ and serves as the basis of our personalized model (see \Cref{fig:transformer_archi}).

\subsection{Pattern Retrieval with a Frozen Foundation Model}

Rather than encoding the full user history as in UserLLM, we retrieve a small set of users' wearable patterns similar to the current input and extract a single vector from them. This enables context-aware personalization under limited data by capturing variability in an individual’s physiological and activity responses, thereby reducing overfitting. To this date, only a few studies have proposed to leverage out-of-domain Foundation model for physiological data, using either Speech Foundation Model (SFM) \cite{phukan_beyond_2024, ahmed_al_dossary_speech_2025}, or TSFM \cite{christenson_assessing_2024}. Here, we experiment with two TSFMs, namely \texttt{MOMENT-1-base} (109M parameters) \cite{goswami_moment_2024}, a T5-based foundation model known for its zero-shot performances in a variety of tasks, \texttt{Chronos-2} (120M parameters) \cite{ansari_chronos-2_2025}, an encoder-only model with time and multivariate time series attention, as well as \texttt{HuBERT-base-ls960} (90M parameters) \cite{hsu_HuBERT_2021}, a popular SFM which here stands as a pure out-of-domain encoder. 

For a given user $u$, we feed each of their samples into the FM, extract embeddings and build a retrieval database $\mathcal{D}^R_u$, with $N_u$ user samples $z^R_u = \mathrm{FM}(x)$. Although Chronos-2 implements group-attention between multiple input time series, the embeddings are of the same size as HuBERT and MOMENT, i.e., $(B, T, M, d_{FM})$ with $B$ being the batch size, $T$ the number of temporal tokens, $M$ the number of modalities, and $d_{FM} = 768$ the dimension of FM's embeddings. In order to have each $B$ samples in the form of a vector, we choose to (a) apply temporal pooling $(B, T, M, d_{FM}) \rightarrow (B, M, d_{FM})$, and (b) concatenate all modality representations $(B, T, M, d_{FM}) \rightarrow (B, d_{FM} \times M)$. The resulting sample embeddings $z^R$ are therefore of size $M \times 768 = 4608$. We provide UMAP visualizations of each foundation model's embeddings, color-coded by label and user, in the supplementary materials. These visualizations reveal that, despite being trained on out-of-domain data, all three models produce embeddings in which samples from the same user form loosely consistent clusters, providing qualitative support for the intra-user cosine similarity retrieval used in our approach. Then given an input $x_{i,u}$ sample from user $u$, top-K elements from the same user are retrieved using cosine similarity from the user's database $\mathcal{D}^R_u$, excluding the corresponding embedding $z^R_{i,u} = \mathrm{FM}(x_{i,u})$. The resulting set of retrieved embeddings form the context set of $x_{i,u}$, denoted as $c_{i,u}$. For simplicity, index $\cdot_{i,u}$ will be dropped in the following section.

\begin{figure}[h!]
    \centering
    \includegraphics[width=.75\linewidth]{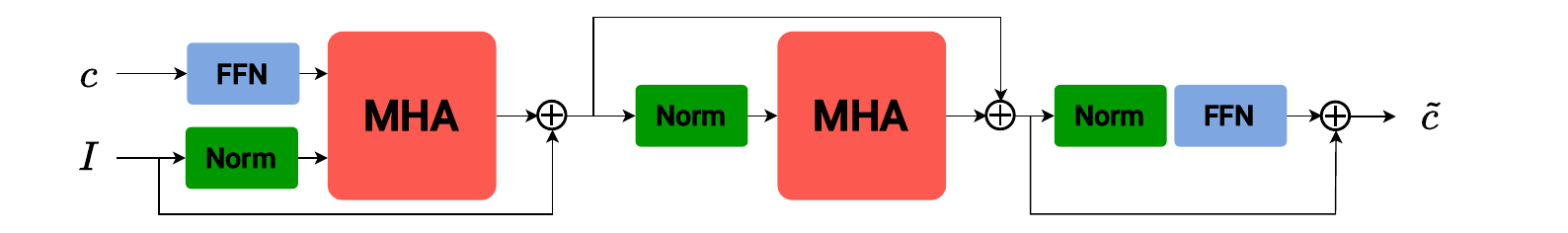}
    \caption{The \textbf{Custom SetTransformer} pools the set $\mathbf{c}$ of $K$ retrieved samples into a smaller set by attending to learned tokens $I$.}
    \label{fig:set_transformer}
\end{figure}

\subsection{Retrieval-augmented personalization}
The set of retrieved pattern $\mathbf{c}$ is then used to condition representation from the main predictive transformer $f(x)$ through a personalized decoder. While UserLLM learns interleaved cross-attention modules in a frozen LLM to condition prediction on user embedding, we choose a much lighter yet efficient approach for conditioning, namely Feature-wise Modulation Layer (FiLM) \cite{perez_film_2018}. We first pool the $K$ retrieved sample into a smaller set with lower dimension elements using a custom and simplified version of the SetTransformer \cite{lee_set_2019} where a set learnable inducing points $I \in \mathbb{R}^{K'\times d_{set}}$, with $K' << K$, are cross-attended with retrieved context $c$ (see \Cref{fig:set_transformer}). This allow us to effectively reduce both the number of elements in the context as well as the retrieved embeddings' dimension. Then, we further reduce the set size to obtain a single condition vector $\bar{c}_i = \mathrm{AvgPool}(\mathbf{c}) \in \mathbb{R}^{d_{set}}$ through average pooling of the $K'$ elements of the reduced set. This conditioning vector is then used as input to a FiLM layer generating the modulating parameters $\gamma$ and $\beta$. We apply such conditioning in a modified self attention block described as:
\begin{align}
    f_{FiLM}(x) &= \gamma f(x) + \beta \\ 
    f_{MHA}(x) &= \mathrm{MHA}(\mathrm{Norm}(f_{FiLM}(x))) + f(x) \\ 
    f_{perso}(x) &= \mathrm{FFN}(\mathrm{Norm}(f_{MHA}(x))) + f_{MHA}(x)
\end{align}
where $f_{perso}(x)$ is the final personalized representation, and FFN, MHA stands for fully connected network and multi-head attention respectively. Personalized prediction is then obtained with a 2-layer FFN with $z_{perso}(x) = \mathrm{MaxPool}(f_{perso}(x)) + \mathrm{MaxPool}(f(x))$ as input.

To prevent non-personalized representation collapse in early stages due to poor performance of the personalized branch, we choose to perform multi-task learning and predict both the base label $\hat{y}_{base}$ from $f(x)$ and the personalized prediction $\hat{y}_{perso}$ from $f_{perso}(x)$. The model is trained on the sum of base and personalized Cross-Entropy losses. In order to dynamically balance these losses, we apply exponential moving average such that the final loss at step $s$ is equal to:

\begin{equation}
    \mathcal{L}_{tot}^s =  \mathrm{CE}(\hat{y}_{base}, y) / \alpha_{base}^s +  \mathrm{CE}(\hat{y}_{perso}, y) / \alpha_{perso}^s  
\end{equation}

where the normalization terms are updated as:
\begin{align}
    \alpha_{perso}^s &= 0.95 \cdot \alpha_{perso}^{s-1} + 0.05 \cdot \mathrm{CE}(\hat{y}_{perso}, y), \\
    \alpha_{base}^s &= 0.95 \cdot \alpha_{base}^{s-1} + 0.05 \cdot \mathrm{CE}(\hat{y}_{base}, y)
\end{align}

These terms act as adaptive scaling factors for each loss at every gradient step (i.e., each batch). For instance, if $\mathrm{CE}(\hat{y}_{perso}, y)$ increases sharply at a given step, $\alpha_{perso}^s$ only partially reflects this increase due to the momentum term (0.95), effectively reducing the relative contribution of the personalized branch. This ensures that, during early training when the personalized branch is unstable, its influence is automatically dampened, thereby preserving the base representation.

\subsection{Datasets}
In this work, we focus on wearable-based data, collecting physiological and accelerometer data, applied to stress. We use the WESAD dataset \cite{schmidt_introducing_2018} which consists of wrist- and chest-worn physiological and activity data of $N=15$ subjects who go through various stressful or calming situations during a lab study. For each user, we extract electro-dermal activity (EDA), temperature (TEMP), blood volume pressure (BVP), and accelerometer (ACC).  In the original WESAD study, windows with a stride of 0.3 seconds are extracted. To avoid redundant sampling in the FM embedding space while preserving enough information for encoding, we extract 60-second windows using a higher stride of 3 seconds. We then select windows with only one corresponding signal, either baseline (0), stress (1), or amusement (2). We used code from the \texttt{dl-4-tsc} Github repository \footnote{https://github.com/Emognition/dl-4-tsc} for data extraction and modified it to perform our pre-processing. We employ similar preprocessing as in \cite{wu_transformer-based_2024}, applying a low-pass Butterworth filter with a cutoff frequency of 0.5 Hz to the EDA signals, and a similar one with a cutoff frequency of 2 Hz for the BVP signal. For the accelerometry, we apply a low-pass filter with cutoff frequency of 10 Hz. Each user's signals are then standardized using z-score normalization, and downsampled to 4 Hz. After extracting windows of 60 seconds with 3 seconds stride, we obtain 11044 samples, with a label distribution of 54 / 30 / 16 for baseline, stress, and amusement respectively. Each user's dataset comprises 736 samples on average and label distributions are consistent with the full dataset.

Along with the main intra-user context-aware personalization, we explore cross-dataset retrieval by using the K-EmoCon dataset \cite{park_k-emocon_2020-1} as our retrieval database $\mathcal{D}_u^R = \mathrm{FM}(\mathcal{D}_{KemoCon})$. K-Emocon is a multimodal dataset of naturalistic conversations, with comprehensive annotations of continuous emotions.  The dataset contains multimodal measurements, including audiovisual recordings, EEG, and peripheral physiological signals, acquired with off-the-shelf devices from 16 sessions of approximately 10-minute long paired debates on a social issue. Similarly to WESAD, the K-EmoCon dataset utilizes the Empatica E4 \footnote{https://www.empatica.com/en-eu/research/e4/}. We build on the original preprocessing code \footnote{https://github.com/Kaist-ICLab/K-EmoCon\_SupplementaryCodes} to remove baseline calibration and extract debate sections and the corresponding wearable signals. We then adopt the same filtering and standardization as above.

\subsection{Evaluation}
In order to evaluate general performance of our approach in a robust way, we perform a repeated Leave-One-Subject-Out Cross-Validation (LOSO CV) with S = 5 different random seeds. We present three different analysis to (a) analyze overall performances with respect to baselines, (b) assess cross-user capabilities in cold start scenarios, and (c) investigate cross-dataset retrieval. 

\paragraph{General performances}
We first report average and standard deviation of accuracy and macro F1-score across the 5 random seeds. To further validate our model's ability to guarantee uniform improvement for most users, we report the median of per user difference \texttt{Median}($\Delta$)  in \% of accuracy between the Transformer baseline and our personalized model. Looking at this median of differences rather than the difference of average between the baseline and the personalized model helps us identify if the personalization is beneficial for all users or not. We also report the personalization ratio \% corresponding to the percentage of users who got their performance increased with personalization out of $N = 15$ individuals w.r.t the Transformer baseline. Finally, to ensure the efficiency of our method, we perform a Linear Mixed Model (LMM) analysis on the accuracy metric to compare our model to baselines, accounting for both random effects (seed) and dependent fixed effects (users), and verify that the Gaussian assumption is satisfied. 

\paragraph{Cold-start} Personalization at deployment requires adapting to new users with limited or no prior data, commonly referred to as the cold-start problem. Our model retrieves intra-user patterns to condition predictions, which requires unlabeled samples from the target user and a foundation model to compute their embeddings. We investigate two retrieval configurations that simulate the cold-start problem under realistic deployment constraints. In the first setting, we simulate a user whose history grows over time by applying a chronological mask to target user data, retaining only samples recorded prior to each query window. The resulting target user embeddings are merged with training user embeddings to form a hybrid retrieval database, progressively shifting from inter-user to intra-user retrieval as more data accumulates. We refer to this setting as \textit{hybrid retrieval}. In the second setting, no target user history is available and retrieval relies exclusively on pre-stored embeddings from training users. Beyond simulating the absence of target user data, this setting also addresses a practical constraint: by pre-storing training user embeddings, the foundation model need not run on-device, reducing memory and privacy overhead. We denote this configuration as \textit{inter user retrieval}. We report average  for all personalized model in both cold start settings in \Cref{tab:cold-start_res}.

\paragraph{Cross-dataset retrieval} We further investigate cross-dataset retrieval, where the retrieval database is populated with embeddings from a different dataset rather than the target user's own history. Specifically, query embeddings are computed from WESAD samples as usual, but the retrieval database consists exclusively of K-EmoCon physiological recordings. When a new sample arrives, its foundation model embedding is matched against K-EmoCon embeddings, and the top-K most similar are pooled via the SetTransformer to form the conditioning vector, following the same pipeline as intra-user retrieval. While both datasets share the same sensing device and physiological modalities, their underlying contexts differ. WESAD involves passive participants exposed to controlled stress stimuli, characterized by phase-dependent, low-frequency trends due to controlled stress stimulus. On the other hand, K-EmoCon captures active individuals engaged in naturalistic debates with continuous arousal and valence annotations, with less predictable physiological responses. This cross-dataset retrieval setting therefore evaluates whether foundation model embeddings can capture physiological patterns that generalize across datasets despite differences in affective context and sensing conditions. We report cross-dataset results in \Cref{tab:cross-dataset_res}. Along with accuracy and macro F1-score, we report the mean normalized concentration index across users, denoted as $C_u$. This is computed as $(H_{max} - H_u) / H_{max}$, where $H_{max} = \log N$ is the maximum entropy under uniform retrieval across the $N=26$ K-EmoCon subjects, and $H_u$ is the empirical entropy of user $u$'s retrieval distribution over those subjects. A value of 0 indicates that retrieval is uniformly distributed across all subjects in the pool, while a value of 1 indicates that all retrieved samples originate from a single subject.

\subsection{Baselines}

We first compare our model to fully supervised approaches, namely a Random Forest with statistical functionals as inputs, and a baseline Transformer, depicted in \Cref{fig:transformer_archi}. To assess the representational quality of out-of-domain foundation models without task-specific adaptation, we perform classification using a simple RBF-kernel SVM on frozen foundation model embeddings, following the linear probing protocol proposed in \cite{goswami_moment_2024}. The objective is not to maximize classification performance through approaches such as parameter-efficient fine-tuning, but rather to evaluate the intrinsic quality of representations produced by each foundation model. Finally, we report results for a fine-tuned Transformer baseline using 1\% and 5\% of each target user's data. Unlike data-level personalization approaches \cite{li_survey_2023-1} that incorporate target user samples directly into the training set, we first train the model on the N-1 training users and then fine-tune the whole model, including the CNN encoder and the transformer, on a small subset of the held-out user's samples, as in \cite{behinaein_transformer_2021}. This simulates personalization at deployment time for a model pre-trained on users it has never encountered. The model is fine-tuned for 5 epochs with a learning rate of 1e-5, a batch size of 16, and no weight decay.

\begin{table*}[t]
\centering
\caption{\textbf{Main results on WESAD} are reported with average and standard deviation of LOSO metrics across 5 random seeds. Along with Accuracy and Macro F1, personalization metrics are reported for \textit{Fine-tuned} and \textit{Retrieval-based Personalized} models (w.r.t to the baseline Transformer).}

\label{tab:main_res}

\begin{tabular}{llcccc}
\toprule
\textbf{Category} & \textbf{Model} & \textbf{Accuracy} & \textbf{Macro F1} & $\mathrm{Median}(\Delta)$ & $\%_{perso}$ \\
\midrule
\textit{Baselines}
& Random Forest                          & 80.73 $\pm$ 0.45 & 75.44 $\pm$ 0.57 & - & - \\
& Baseline Transformer                 & 83.99 $\pm$ 0.89 & 77.07 $\pm$ 1.00 & - & -\\
\midrule
\textit{FM + SVM}
& MOMENT + SVM                           & 77.12 & 67.28 & - & -\\
& Chronos + SVM                          & 81.19 & 73.87 & - & -\\
& HuBERT + SVM                           & 76.70 & 67.66 & - & -\\
\midrule
\textit{Retrieval-based}
& Ours (MOMENT)                          & 87.91 $\pm$ 0.56 & 81.83 $\pm$ 0.84 & 3.21 & 70.67 \\
\textit{Personalization}
& Ours (Chronos)                         & 87.58 $\pm$ 0.68 & 81.52 $\pm$ 1.25 & 3.49 & 78.67\\
& Ours (HuBERT)                          & 86.94 $\pm$ 0.95 & 80.91 $\pm$ 1.07 & 3.30 & 65.33\\
\midrule
\textit{Fine-tuned}
& Fine-tune 1 \%  (\textit{$\sim$ 8 samples})    & 88.08 $\pm$ 2.49 & 84.85 $\pm$ 2.58 & 2.80 & 70.67 \\
& Fine-tune 5 \%  (\textit{$\sim$ 40 samples})    & 96.23 $\pm$ 0.34 & 95.05 $\pm$ 0.75 & 10.13 &97.33 \\

\bottomrule
\end{tabular}%

\end{table*}

\subsection{Implementation details}
\paragraph{CNN+Transformer encoder}
We implement the CNN encoder with 64 filters per modality, yielding concatenated features of size $(L, F \times M) = (240, 64 \times 6 = 384)$. These convolutional features are then fed to a 6-layer Transformer with $d_{model} = 128$, 4 attention heads, and a feedforward dimension of $4 \times d_{model} = 512$.

\paragraph{Foundation model embeddings}
We employ two strategies for extracting foundation models' embeddings. Given that Chronos-2 supports variable-length multivariate input, we therefore feed the downsampled sequence of size $(L, M) = (240, 6)$ directly to the model. For MOMENT and HuBERT, we apply linear interpolation to upsample the wearable signals to the required input length. MOMENT expects sequences of length 512, while HuBERT has no fixed length requirement but is designed for audio sampled at 16kHz, to which we upsample accordingly. Each modality is processed independently by the foundation model, yielding embeddings of size $(L, M, d_{FM}) = (240, 6, 768)$, which are then average-pooled along $L$ and concatenated across modalities to produce a single vector of size $M \times d_{FM} = 4608$ per sample.

\paragraph{Personalization modules}
We retrieve $K = 16$ samples per query, corresponding to approximately 2\% of each user's available history, providing sufficient diversity while limiting noise. The retrieved context is pooled by a custom SetTransformer with hidden dimension $d_{set} = 128$, 4 heads for both self- and cross-attention, and $K' = 4$ inducing points. The resulting average-pooled conditioning vector $\bar{c}$ is projected into a 64-dimensional space before being passed to 2 blocks of FiLM + Self-Attention with the same hidden dimension.

\paragraph{Full model and training}
Both the personalized and base prediction heads consist of a linear layer with ReLU activation mapping to 64 dimensions, followed by a final linear layer producing class logits. To limit overfitting under LOSO, we adopt a compact architecture with aggressive regularization, namely high weight decay and high dropout, combined with learning rate scheduling. Both the baseline Transformer and the retrieval-based personalized model are trained for 20 epochs with a learning rate of $10^{-4}$, a weight decay of $10^{-2}$, and a batch size of 128, with a minimum learning rate of $10^{-6}$. All models were trained on an NVIDIA RTX A6000, with training times ranging from 35 minutes to 1 hour for the full LOSO evaluation depending on the configuration \footnote{Code used for experiments will be made available upon acceptance}.

\section{Results}

\subsection{Main results}
\label{sec:main}
We first analyze the overall performance of our personalized model and report results in \Cref{tab:main_res}. The Random Forest baseline achieves 80.73\% accuracy and 75.44\% F1, while the baseline Transformer improves upon it at 83.99\% accuracy and 77.07\% F1. Linear probing of frozen foundation model embeddings performs below both supervised baselines across all models, with the exception of Chronos which marginally exceeds the Random Forest in accuracy. This confirms that out-of-domain foundation models do not produce representations sufficiently discriminative for direct stress classification, and that naive transfer of these embeddings to physiological affect recognition is insufficient. 

On the other hand, all three variants of our retrieval-based personalization approach consistently outperform the non-personalized baseline Transformer, with MOMENT achieving the highest accuracy (87.91\% $\pm$ 0.56\%) and macro F1-score (81.83\% $\pm$ 0.84\%), corresponding to gains of +3.92\% and +4.76\% respectively. The LMM analysis confirms that these gains are statistically significant ($\beta$ = 3.915, 95\% CI [2.590, 5.241],  z = 5.789, p < 0.001 for MOMENT, $\beta$ = 3.490,  95\% CI [2.402, 4.578],  z = 6.287, p < 0.001 for Chronos, and $\beta$ = 2.950,  95\% CI [1.776, 4.124],  z = 4.926, p < 0.001). When it comes to personalization gain, Chronos achieves the highest median per-user gain (Median($\Delta$) = 3.49) and win rate (78.67\%), suggesting more consistent improvements across individuals than MOMENT despite slightly lower aggregate performance. Finally, our method approaches the performance of supervised fine-tuning with 1\% of target user data (88.08\% / 84.85\%) without requiring any labeled user samples, closing most of the gap between the non-personalized baseline and the fine-tuned model. It remains however well below 5\% fine-tune, indicating that our model still fails at capturing some users' specificity. A detailed per-user breakdown of accuracy and macro F1-score for baseline Transformer and our proposed model is provided in the supplementary materials.

\subsection{Cold-start}
\label{sec:cold-start_res}

\begin{figure}[h!]
    \centering
    \includegraphics[width=.75\linewidth]{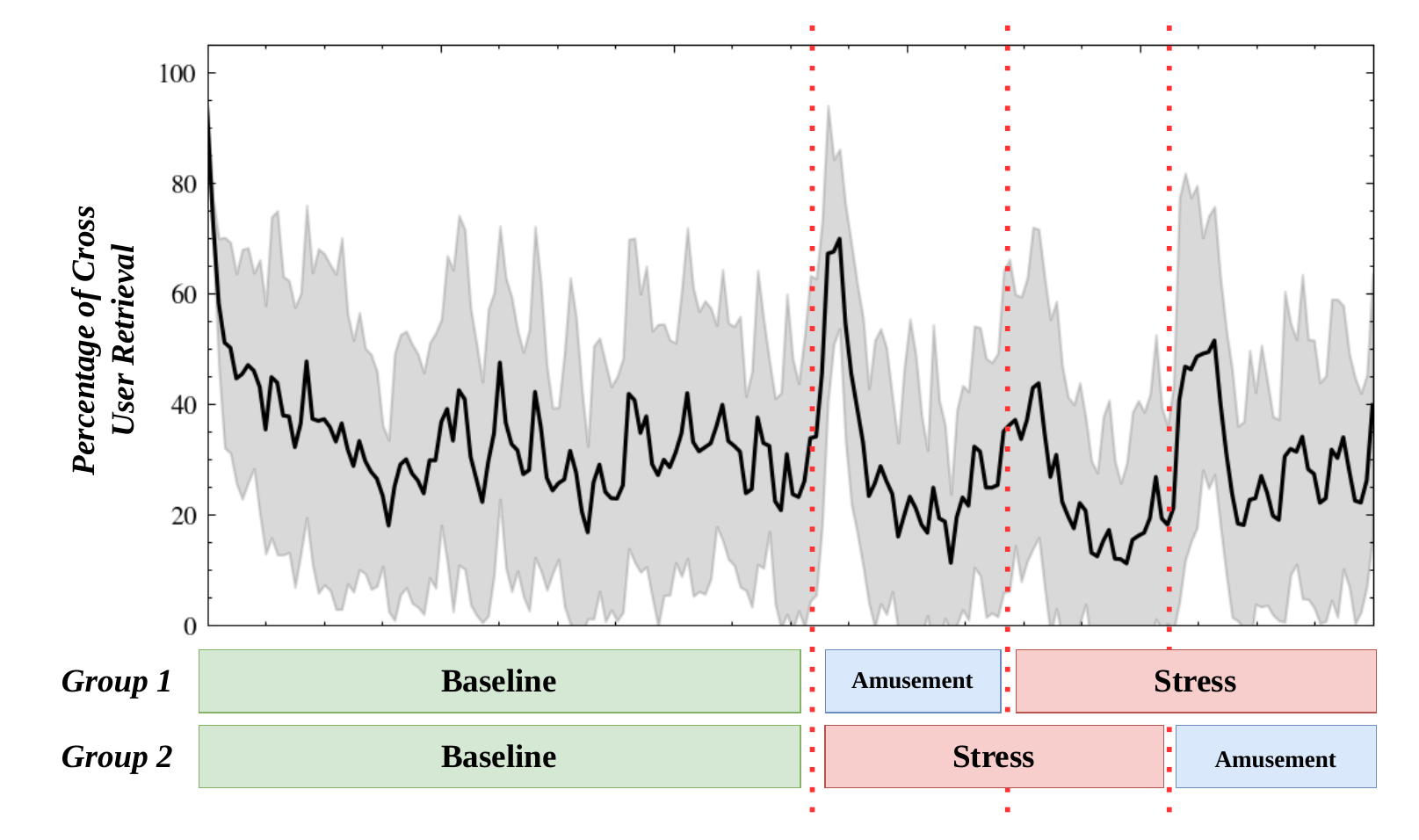}
    \caption{The temporal evolution of the cross-user retrieval ratio in hybrid cold-start.}
    \label{fig:temporal_cross-user}
\end{figure}

We report in \Cref{tab:cold-start_res} results under both cold-start configurations. The hybrid retrieval setting, which combines chronologically prior samples from the target user with training user embeddings, achieves performance close to full intra-user retrieval (see \Cref{tab:main_res}) across all foundation models (MOMENT: 87.27\% accuracy, 81.20\% F1; Chronos: 87.04\% / 81.33\%; HuBERT: 85.33\% / 78.92\%). This demonstrates that even a partial, temporally ordered user history is sufficient for effective personalization. On the other hand, inter-user retrieval, i.e.,  where no target user data is available and retrieval relies exclusively on training user embeddings, yields substantially lower performance across all models (MOMENT: 78.90\% / 63.23\%; Chronos: 78.79\% / 68.54\%; HuBERT: 77.92\% / 69.04\%), falling below the non-personalized baseline. This suggests that the personalization gain partially stems from user-specific physiological patterns rather than generalized, user-invariant representations of wearable signals, validating our intra-user retrieval approach.

\begin{table}[h!]
\centering
\caption{\textbf{Cold-start results} in hybrid and inter-user only settings.}
\label{tab:cold-start_res}
\begin{tabular}{llcc}
\toprule
\textbf{Mode} & \textbf{Model} & \textbf{Accuracy} & \textbf{Macro F1} \\
\midrule
\textit{Hybrid} & Moment & 87.27 & 81.20 \\
& Chronos & 87.04  & 81.33 \\
& HuBERT & 85.33 & 78.92  \\
\midrule
\textit{Inter-user} & Moment & 78.90 & 63.23  \\
& Chronos & 78.79 & 68.54  \\
& HuBERT & 77.92 & 69.04  \\
\bottomrule
\end{tabular}
\end{table} 

To further analyze hybrid cold-start, we represent in \Cref{fig:temporal_cross-user} the temporal evolution of the cross-user retrieval ratio, averaged across all users and seeds with $\pm 1$ std. At the beginning of each session, the temporal mask ensures that all retrieved embeddings originate from other users, yielding a cross-user ratio of 100\%. This ratio decreases to approximately 30\% as the target user accumulates history, with a few notable peaks corresponding to transitions between protocol phases. The WESAD protocol alternates between baseline, amusement, and stress phases, with the order of the last two permuted across participants. We represent this in the figure with Group 1 $\rightarrow$ baseline, amusement, stress and  Group 2 $\rightarrow$ baseline, stress, amusement. At each phase transition, the target user has no prior embeddings from the new phase, causing the model to temporarily fall back on other users' embeddings, with cross-user retrieval rising to approximately 70\% on average at the first transition. This behavior suggests that the model adaptively leverages inter-user context when intra-user history is insufficient, rather than relying on potentially irrelevant recent samples from a different phase. The first transition yields the highest peak, which we attribute to the frequent confusion between baseline and amusement phases observed in our predictions, as these two states are the most physiologically similar in the WESAD protocol.

\subsection{Cross-dataset retrieval}

\begin{table}[h!]

\caption{\textbf{Cross-dataset retrieval} results on the K-EmoCon dataset.}
\centering
\begin{tabular}{lccc}
\toprule
\textbf{Model} & \textbf{Accuracy} & \textbf{Macro F1} & \textbf{$C_u$}(\%) \\
\midrule
Moment & 83.50 & 76.41 & 22.13 \\ 
Chronos & 83.29 & 76.40 & 9.26 \\ 
HuBERT & 83.39 & 76.45 &  14.36 \\ 
\bottomrule
\end{tabular}
\label{tab:cross-dataset_res}
\end{table} 

Cross-dataset retrieval results are reported in \Cref{tab:cross-dataset_res}, where the retrieval database exclusively contains K-EmoCon embeddings. All three foundation models achieve performance close to the non-personalized baseline Transformer, which is notably higher than inter-user retrieval reported in \Cref{tab:cold-start_res}. This suggests that K-EmoCon physiological recordings, despite originating from a different affective context, provide more useful retrieval context than other WESAD users' embeddings. However, the mean normalized concentration index $C_u$ remains low across users, indicating that retrieval is broadly distributed across K-EmoCon subjects rather than concentrating on a few physiologically similar ones. Drawing on our observations in \Cref{sec:cold-start_res}, where consistent intra-user retrieval was shown to be key for effective personalization, the low concentration index suggests that the model cannot identify sufficiently consistent cross-dataset patterns to build a meaningful conditioning embedding. We hypothesize that in this setting, the model may be discarding personalized context via the residual connection in the personalized branch, effectively falling back on the base representation.

To verify this, we measure the personalization magnitude, defined as the normalized difference between the base and personalized representations in the final prediction: $||\Delta z||_2 / ||z_{base}||_2$, where $z_{base} = \mathrm{MaxPool}(f(x))$ and $\Delta z = z_{base} - \mathrm{MaxPool}(f_{perso}(x))$. A value close to zero indicates that the personalized branch contributes little beyond the base representation.

\begin{figure}[h!]
    \centering
    \includegraphics[width=.75\linewidth]{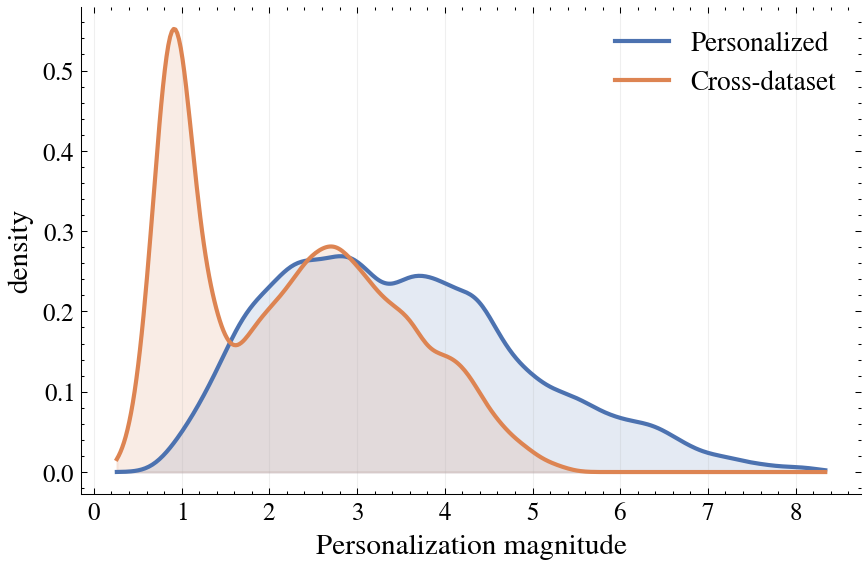}
    \caption{\textbf{Personalization magnitude} measures the contribution of personalized representation in the final prediction (the closer to zero, the weaker the personalization).}
    \label{fig:residual-analysis}
\end{figure}

\Cref{fig:residual-analysis} shows the distribution of this quantity for the full intra-user model and the cross-dataset variant, both using MOMENT. The intra-user model exhibits a broad, high-magnitude distribution, reflecting active personalization across users and labels. In contrast, the cross-dataset model shows a sharp peak at low magnitude, confirming that the model suppresses the personalized context when it originates from K-EmoCon, relying predominantly on the base representation. This may indicate why cross-dataset performance approaches the non-personalized baseline rather than full intra-user personalization.

\subsection{Ablation studies}

\begin{table}[h]
\centering
\caption{\textbf{Ablation study}: Results are averaged over 5 seeds. The $^\dagger$non-personalized baseline  and our $^*$personalized model (MOMENT version) constitute lower and upper bound respectively.}
\label{tab:ablation}
\begin{tabular}{lcccc}
\toprule
\textbf{Model} & \textbf{Accuracy} & \textbf{Macro F1} \\
\midrule
\textit{Full personalized model} & & \\
\quad MOMENT + SetTransformer + FiLM $^*$ & 87.91 & 81.83 \\
\midrule
\textit{Conditioning} & & \\
\quad MOMENT + SetTransformer + Concat. & 87.33 & 81.21  \\
\midrule
\textit{Retrieval backbone} & & \\
\quad VAE + SetTransformer + FiLM & 81.48 & 75.28  \\
\midrule
\textit{No personalization} & & \\
\quad baseline Transformer$^\dagger$ & 83.99 & 77.07  \\
\bottomrule
\end{tabular}%

\end{table}

Table~\ref{tab:ablation} reports ablation results averaged over 5 seeds. Replacing the foundation model retrieval backbone with a VAE trained on WESAD physiological signals substantially degrades performance (81.48\% accuracy, 75.28\% F1), falling below the non-personalized baseline (83.99\% / 77.07\%). This confirms that the pretrained out-of-domain foundation model's representations are essential for meaningful retrieval. A domain-specific unsupervised model such as a VAE trained on limited physiological data does not provide sufficient embedding quality to support effective personalization. Replacing FiLM conditioning with simple concatenation yields a modest reduction (87.33\% / 81.21\% vs. 87.91\% / 81.83\%), suggesting that our method is robust to the choice of conditioning mechanism, and that the primary source of personalization gain lies in the quality of the retrieved embeddings rather than the conditioning architecture.

\section{Conclusion}
In this study, we introduced a personalization approach that retrieves similar patterns from a user's own history using frozen, out-of-domain foundation models to condition a lightweight stress detection model. Our method, trained end-to-end without any labeled user data from target users, consistently outperforms a non-personalized baseline and approaches the performance of supervised fine-tuning. We further show that even with limited user history, our method remains effective, and that the model naturally discards uninformative retrieval context when it is not useful. These results suggest that frozen foundation models, despite being trained on unrelated data, can serve as effective retrieval backbones for personalization in domains where large-scale physiological data is scarce. There are, however, few limitations to this work. While our approach eliminates the need for labeled user data, it still requires sufficient unlabeled samples from each user, limiting applicability in true cold-start scenarios. Additionally, despite uniform gains, some users still show modest improvements, and the foundation model's size may constrain deployment in resource-limited settings. Finally, despite exploring cross-dataset retrieval, we evaluated our approach only on one dataset. Future work will explore model compression, investigate other out-of-domain foundation models and their fine-tuning, e.g., through LoRA, and extension to other multimodal affect recognition settings.

\section*{Safe and Responsible Innovation Statement}
In this study, we propose a novel personalization method and evaluate it on two publicly available datasets, namely WESAD and K-EmoCon, collected in laboratory settings with participants' informed consent. No demographic attributes, e.g., age, gender, or ethnicity, were used in order to limit potential bias. However, our proposed method relies on unlabeled physiological data collected from target users, which raises privacy considerations in deployment and calls for explicit informed consent from participants. We emphasize that the proposed method is not intended for clinical or diagnostic use, as the laboratory-controlled conditions used for data collection may not reflect real-world scenarios.

\bibliographystyle{plainnat}
\bibliography{biblio}

\appendix

\begin{figure}[h]
  \centering
  \begin{subtable}[t]{0.48\textwidth}
    \centering
    \begin{tabular}{lcc}
      \toprule
      \textbf{User} & \textbf{Accuracy} & \textbf{Macro F1} \\
      \midrule
      2 & $79.18 \pm 2.87$ & $69.24 \pm 2.78$ \\
      3 & $93.05 \pm 1.57$ & $92.34 \pm 2.04$ \\
      4 & $88.01 \pm 1.17$ & $85.21 \pm 1.47$ \\
      5 & $87.93 \pm 3.12$ & $86.83 \pm 2.93$ \\
      6 & $84.82 \pm 2.17$ & $79.93 \pm 2.99$ \\
      7 & $79.64 \pm 2.64$ & $68.28 \pm 2.62$ \\
      8 & $89.16 \pm 1.43$ & $80.97 \pm 3.43$ \\
      9 & $99.04 \pm 0.29$ & $98.78 \pm 0.35$ \\
      10 & $86.42 \pm 1.81$ & $82.13 \pm 2.34$ \\
      11 & $88.13 \pm 2.88$ & $86.28 \pm 3.11$ \\
      13 & $81.51 \pm 0.83$ & $61.07 \pm 0.73$ \\
      14 & $57.92 \pm 4.06$ & $44.17 \pm 1.80$ \\
      15 & $82.72 \pm 5.29$ & $76.27 \pm 7.29$ \\
      16 & $96.44 \pm 1.64$ & $94.99 \pm 2.30$ \\
      17 & $65.93 \pm 0.38$ & $49.55 \pm 1.55$ \\
      \midrule
      \textbf{Mean} & $83.99 \pm 0.89$ & $77.07 \pm 1.00$ \\
      \bottomrule
    \end{tabular}
    \caption{Baseline}
    \label{tab:baseline}
  \end{subtable}
  \hfill
  \begin{subtable}[t]{0.48\textwidth}
    \centering
    \begin{tabular}{lcc}
      \toprule
      \textbf{User} & \textbf{Accuracy} & \textbf{Macro F1} \\
      \midrule
      2 & \cellcolor{green!20}$94.00 \pm 1.67$ & \cellcolor{green!20}$92.35 \pm 2.17$ \\
      3 & $93.42 \pm 2.79$ & $92.19 \pm 2.68$ \\
      4 & $87.65 \pm 2.76$ & $80.48 \pm 5.55$ \\
      5 & $91.66 \pm 2.17$ & $90.37 \pm 2.46$ \\
      6 & \cellcolor{green!20}$92.26 \pm 1.73$ & \cellcolor{green!20}$89.00 \pm 2.36$ \\
      7 & \cellcolor{green!20}$93.77 \pm 1.68$ & \cellcolor{green!20}$92.05 \pm 2.05$ \\
      8 & $89.86 \pm 1.44$ & $83.43 \pm 2.72$ \\
      9 & $99.97 \pm 0.05$ & $99.96 \pm 0.07$ \\
      10 & $86.66 \pm 2.66$ & $78.16 \pm 3.83$ \\
      11 & \cellcolor{green!20}$98.03 \pm 0.77$ & \cellcolor{green!20}$97.43 \pm 0.97$ \\
      13 & $77.76 \pm 1.88$ & $60.84 \pm 1.30$ \\
      14 & \cellcolor{green!20}$70.41 \pm 3.08$ & \cellcolor{green!20}$53.77 \pm 2.09$ \\
      15 & $83.49 \pm 2.85$ & $76.85 \pm 3.73$ \\
      16 & \cellcolor{red!20}$89.85 \pm 3.12$ & \cellcolor{red!20}$82.49 \pm 5.38$ \\
      17 & \cellcolor{green!20}$69.83 \pm 2.19$ & \cellcolor{green!20}$58.13 \pm 3.65$ \\
      \midrule
      \textbf{Mean} & $87.91 \pm 0.56$ & $81.83 \pm 0.84$ \\
      \bottomrule
    \end{tabular}
    \caption{Personalised (Moment)}
    \label{tab:mainperso_moment}
  \end{subtable}
  \\[1em]
  \begin{subtable}[t]{0.48\textwidth}
    \centering
    \begin{tabular}{lcc}
      \toprule
      \textbf{User} & \textbf{Accuracy} & \textbf{Macro F1} \\
      \midrule
      2 & \cellcolor{green!20}$88.49 \pm 1.83$ & \cellcolor{green!20}$83.76 \pm 2.90$ \\
      3 & $95.62 \pm 1.63$ & $94.89 \pm 1.85$ \\
      4 & $90.03 \pm 3.45$ & $84.27 \pm 6.27$ \\
      5 & $90.64 \pm 2.87$ & $89.27 \pm 2.99$ \\
      6 & \cellcolor{green!20}$91.39 \pm 1.78$ & \cellcolor{green!20}$87.11 \pm 2.87$ \\
      7 & \cellcolor{green!20}$91.53 \pm 1.49$ & \cellcolor{green!20}$87.83 \pm 2.93$ \\
      8 & $89.13 \pm 1.25$ & $80.85 \pm 3.08$ \\
      9 & $96.67 \pm 0.77$ & $96.14 \pm 0.82$ \\
      10 & $85.28 \pm 1.23$ & $79.48 \pm 1.58$ \\
      11 & \cellcolor{green!20}$94.89 \pm 2.61$ & \cellcolor{green!20}$93.29 \pm 3.22$ \\
      13 & $80.67 \pm 3.14$ & \cellcolor{green!20}$71.55 \pm 4.32$ \\
      14 & \cellcolor{green!20}$65.00 \pm 1.85$ & $46.53 \pm 1.50$ \\
      15 & $82.23 \pm 4.40$ & \cellcolor{red!20}$69.14 \pm 6.79$ \\
      16 & \cellcolor{red!20}$92.28 \pm 5.94$ & $91.00 \pm 6.30$ \\
      17 & \cellcolor{green!20}$70.30 \pm 2.93$ & \cellcolor{green!20}$58.51 \pm 5.27$ \\
      \midrule
      \textbf{Mean} & $86.94 \pm 0.95$ & $80.91 \pm 1.07$ \\
      \bottomrule
    \end{tabular}
    \caption{Personalised (HuBERT)}
    \label{tab:mainperso_hubert}
  \end{subtable}
  \hfill
  \begin{subtable}[t]{0.48\textwidth}
    \centering
    \begin{tabular}{lcc}
      \toprule
      \textbf{User} & \textbf{Accuracy} & \textbf{Macro F1} \\
      \midrule
      2 & \cellcolor{green!20}$87.55 \pm 1.37$ & \cellcolor{green!20}$82.23 \pm 2.18$ \\
      3 & $96.60 \pm 0.39$ & $95.61 \pm 0.69$ \\
      4 & \cellcolor{green!20}$92.05 \pm 2.79$ & \cellcolor{green!20}$89.28 \pm 4.31$ \\
      5 & $88.77 \pm 2.21$ & $87.36 \pm 1.94$ \\
      6 & $87.87 \pm 0.46$ & $80.26 \pm 1.20$ \\
      7 & \cellcolor{green!20}$89.23 \pm 3.03$ & \cellcolor{green!20}$83.93 \pm 4.54$ \\
      8 & $90.68 \pm 0.52$ & \cellcolor{green!20}$84.87 \pm 1.34$ \\
      9 & $99.89 \pm 0.13$ & $99.86 \pm 0.17$ \\
      10 & $88.69 \pm 2.91$ & $83.14 \pm 5.40$ \\
      11 & \cellcolor{green!20}$97.77 \pm 1.84$ & \cellcolor{green!20}$97.14 \pm 2.25$ \\
      13 & \cellcolor{red!20}$77.60 \pm 2.68$ & $62.06 \pm 2.83$ \\
      14 & \cellcolor{green!20}$65.62 \pm 1.61$ & $47.82 \pm 1.84$ \\
      15 & $82.42 \pm 4.37$ & $73.66 \pm 6.72$ \\
      16 & $94.30 \pm 1.67$ & \cellcolor{red!20}$91.26 \pm 2.83$ \\
      17 & \cellcolor{green!20}$73.20 \pm 4.80$ & \cellcolor{green!20}$64.27 \pm 7.69$ \\
      \midrule
      \textbf{Mean} & $87.48 \pm 0.66$ & $81.52 \pm 1.25$ \\
      \bottomrule
    \end{tabular}
    \caption{Personalised (Chronos)}
    \label{tab:mainperso_chronos}
  \end{subtable}
  \caption{\textbf{Per-user accuracy and macro F1} (mean $\pm$ std across seeds). Cells highlighted in \colorbox{green!20}{green} (resp.\ \colorbox{red!20}{red}) indicate improvements (resp.\ degradations) exceeding the median absolute difference with respect to the baseline.}
  \label{fig:perf_tables}
\end{figure}

\newpage

\begin{figure}[t]
  \centering
  \includegraphics[width=\linewidth]{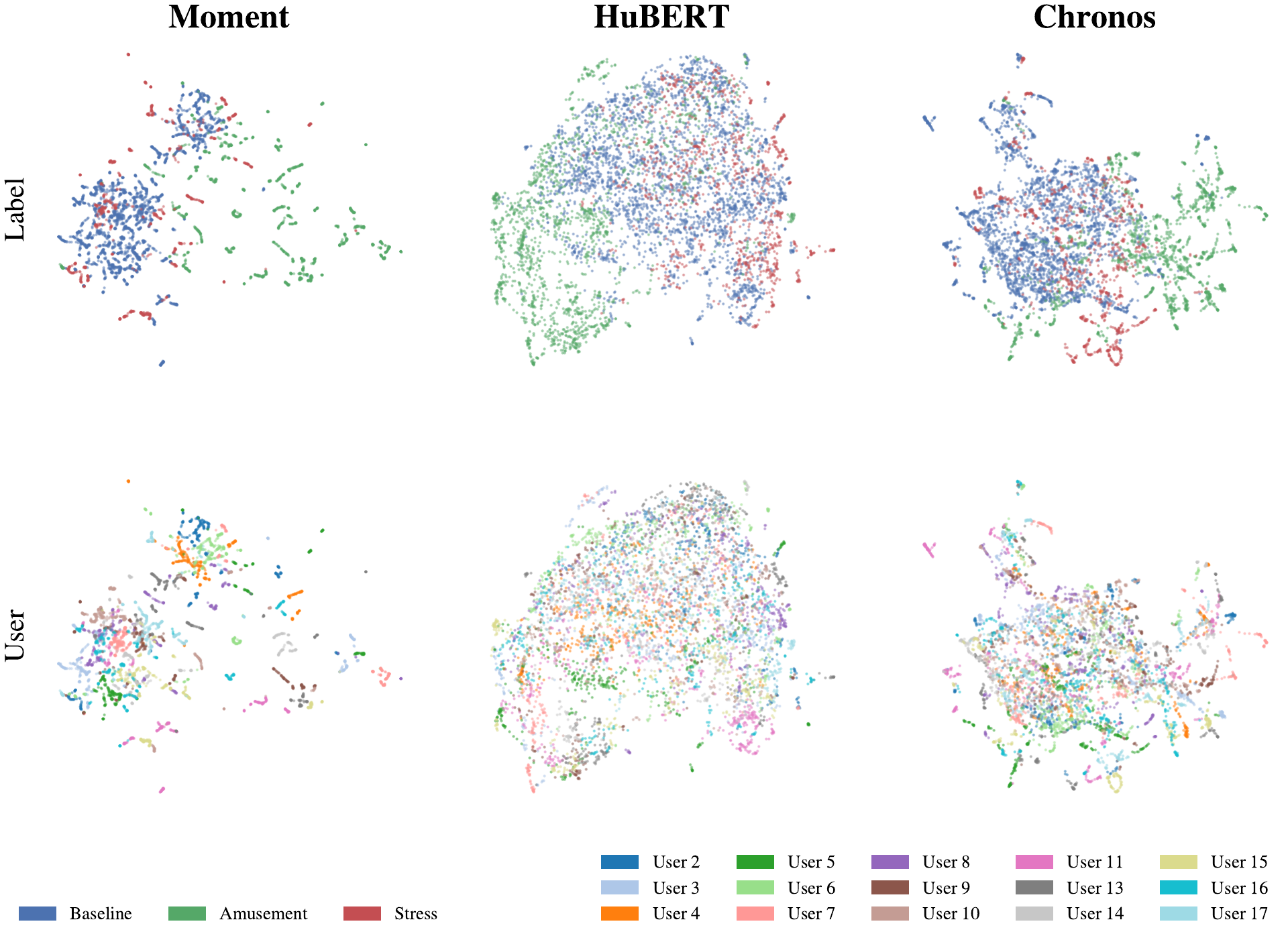}
  \caption{%
    UMAP projections of raw foundation-model embeddings on WESAD.
    \textbf{Top row}: colored by stress label
    (\textcolor{blue}{baseline}, \textcolor{green!60!black}{amusement}, \textcolor{red}{stress}).
    \textbf{Bottom row}: colored by participant.
  }
  \label{fig:fm_umap}
\end{figure}

\end{document}